%% file: main.tex
\documentclass{article}

\usepackage{microtype}
\usepackage{graphicx}
\usepackage{subfigure}
\usepackage{booktabs} 

\usepackage{hyperref}

\usepackage[accepted]{icml2025}

\usepackage{amsmath}
\usepackage{amssymb}
\usepackage{mathtools}
\usepackage{amsthm}

\usepackage[capitalize,noabbrev]{cleveref}

\usepackage{color}

\usepackage{xspace}
\newcommand{\name}[0]{SwapGT\xspace}

\makeatletter
\DeclareRobustCommand\onedot{\futurelet\@let@token\@onedot}
\def\@onedot{\ifx\@let@token.\else.\null\fi\xspace}

\makeatother

\theoremstyle{plain}

\theoremstyle{definition}

\theoremstyle{remark}

\usepackage[textsize=tiny]{todonotes}

\icmltitlerunning{Rethinking Tokenized Graph Transformers for Node Classification}

\begin{document}

\twocolumn[
\icmltitle{Rethinking Tokenized Graph Transformers for Node Classification}


\icmlsetsymbol{equal}{*}

\begin{icmlauthorlist}
\icmlauthor{Jinsong Chen}{equal,cs,hccs}
\icmlauthor{Chenyang Li}{equal,cs,hccs}
\icmlauthor{GaiChao Li}{cs,hccs}
\icmlauthor{John E. Hopcroft}{hccs,cornell}
\icmlauthor{Kun He}{cs,hccs}
\end{icmlauthorlist}

\icmlaffiliation{cs}{School of Computer Science and Technology, Huazhong University of Science and Technology, China.}
\icmlaffiliation{hccs}{Hopcroft Center on Computing Science, Huazhong University of Science and Technology,  China}
\icmlaffiliation{cornell}{Department of Computer Science, Cornell University, USA}

\icmlcorrespondingauthor{Kun He}{brooklet60@hust.edu.cn}


\vskip 0.3in
]



\printAffiliationsAndNotice{\icmlEqualContribution} 

\begin{abstract}
Node tokenized graph Transformers (GTs) have shown promising performance in node classification. The generation of token sequences is the key module in existing tokenized GTs which transforms the input graph into token sequences, facilitating the node representation learning via Transformer. In this paper, we observe that the generations of token sequences in existing GTs only focus on the first-order neighbors on the constructed similarity graphs, which leads to the limited usage of nodes to generate diverse token sequences, further restricting the potential of tokenized GTs for node classification. To this end, we propose a new method termed SwapGT. SwapGT first introduces a novel token swapping operation based on the characteristics of token sequences that fully leverages the semantic relevance of nodes to generate more informative token sequences. Then, SwapGT leverages a Transformer-based backbone to learn node representations from the generated token sequences. Moreover, SwapGT develops a center alignment loss to constrain the representation learning from multiple token sequences, further enhancing the model performance. Extensive empirical results on various datasets showcase the superiority of SwapGT for node classification.

\end{abstract}



\section{Introduction}
\label{Sec:intro}
\input{Body/0Intro}

\section{Relation Work}
\label{Sec:rw}

\input{Body/1RW}

\section{Preliminaries}
\label{Sec:pre}

\input{Body/2Pre}

\section{Methodology}
\label{Sec:method}

\input{Body/3Method}

\section{Experiments}
\label{Sec:exp}
\input{Body/4Exp}

\section{Conclusion}
\label{Sec:con}
In this paper, we introduced a novel tokenized Graph Transformer \name for node classification.
In \name, we developed a novel token swapping operation that flexibly swaps tokens in different token sequences, thereby generating diverse token sequences. This enhances the model’s ability to capture rich node representations.  
Furthermore, \name employs a tailored Transformer-based backbone with a center alignment loss to learn node representations from the generated  multiple token sequences. The center alignment loss helps guide the learning process when nodes are associated with multiple token sequences, ensuring that the learned representations are consistent and informative. 
Experimental results demonstrate that \name significantly improves node classification performance, outperforming several representative GT and GNN models.






\section*{Impact Statement}
This paper presents work whose goal is to advance the field of graph representation learning. There is none potential societal consequence of our work that must be specifically highlighted here.





\bibliography{reference}
\bibliographystyle{icml2025}

\newpage
\appendix
\onecolumn

\input{Body/5APP}




\end{document}

%% file: Body/0Intro.tex
Node classification, the task of predicting node labels in a graph, is a fundamental  problem in graph data mining with numerous real-world applications. 
Graph Neural Networks (GNNs)~\cite{gnn1,gcn} have traditionally been the dominant approaches. 
However, the message passing mechanism inherent to GNNs suffers from some limitations, such as over-smoothing~\cite{oversm}, 
which prevents them from effectively capturing deep graph structural information and hinders their performance in downstream tasks.

In contrast, Graph Transformers (GTs), which adapt the Transformer framework for graph-based learning, have emerged as a promising alternative, demonstrating impressive performance in node classification. Existing GTs can be broadly classified into two categories based on their model architecture: hybrid GTs and tokenized GTs. 

Hybrid GTs combine the strengths of GNN and Transformer, using GNNs to capture local graph topology and Transformers to model global semantic relationships. 
However, recent studies have highlighted a key issue with this approach: 
directly modeling semantic correlations among all node pairs using Transformers can lead to the over-globalization problem~\cite{cob}, which compromises model performance. 

Tokenized GTs, on the other hand, generate independent token sequences for each node, which encapsulate both local topological and global semantic information.
Transformer models are then utilized to learn node representations from these token sequences. 
The advantage of tokenized GTs is that they limit the token sequences to a small, manageable number of tokens, naturally avoiding the over-globalization issue. 
In tokenized GTs, the token sequences typically include two types of tokens: neighborhood tokens and node tokens. Neighborhood tokens aggregate multi-hop neighborhood information of a target node, while node tokens are sampled based on the similarity between nodes.

However, recent studies~\cite{vcrgt} have shown that neighborhood tokens often fail to preserve complex graph properties such as long-range dependencies and heterophily, limiting the richness of node representations. On the other hand, node tokens, generated through various sampling strategies, can better capture correlations between nodes in both feature and topological spaces~\cite{ansgt,ntformer}, making them more effective in preserving complex graph information. As a result, this paper focuses on node token-based GTs.

A recent study~\cite{ntformer} formalized the node token generation process as two key steps: similarity evaluation and top-$k$ sampling. 
In the first step, similarity scores between node pairs are calculated based on different similarity measures to preserve the relations of nodes in different feature spaces. While in the second step, the top $k$ nodes with the highest similarity scores are selected as node tokens to construct the token sequence. 
In this paper, we provide a new perspective on token generation in existing tokenized GTs. 
We identify that the token generation process can be viewed as a neighbor selection operation on the $k$-nearest neighbor ($k$-NN) graph.  
Specifically, a $k$-NN graph is constructed based on node pair similarities, and the neighbor nodes within the first-order neighborhood of each node are selected to form  the token sequence.

\begin{figure}[h]
    \centering    
    \includegraphics[width=7.5cm]{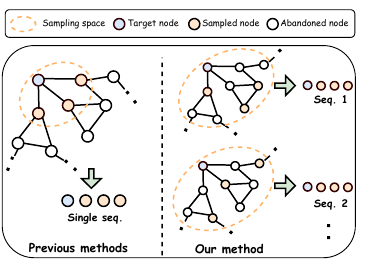}
    \caption{The toy example of token generation on the $k$-NN graph. Previous methods only focus on 1-hop neighborhood to construct a single token sequence. While our method can flexibly select tokens from multi-hop neighborhoods to generate diverse token sequences.}
    \label{fig:motivation}
\end{figure}

\autoref{fig:motivation} illustrates this idea with a toy example.
We can observe that only a small subset of nodes is selected via existing token generation strategies, which indicates that existing methods have limited exploitation of the $k$-NN graph and are unable to comprehensively utilize the correlations between node pairs to explore more informative nodes with potential association to construct token sequences. 
This situation inevitably restricts the ability of tokenized GTs to capture informative node representations. 
Furthermore, in scenarios with sparse training data, relying on token sequences generated from a limited set of nodes may lead to over-fitting, as Transformers, being complex models, may struggle to generalize effectively.

This leads to the following research question: \textit{How can we more comprehensively and effectively exploit node pair correlations to generate diverse token sequences, thus improving the performance of tokenized GTs for node classification?} 
To address this, we introduce a novel method called \name. 
Specifically, \name introduces a new operation, token swapping, which leverages the  semantic correlations of nodes in the $k$-NN to swap tokens in different token sequences, generating more diverse token sequences.
By incorporating multiple token sequences, \name 
enables the model to learn more comprehensive node representations. 
Additionally, \name employs a Transformer-based backbone and introduces a tailored readout function to learn node representations from the generated token sequences.
To handle the case where a node is assigned multiple token sequences, we propose a center alignment loss to guide the training process. 
The main contributions of this paper are summarized as follows:
\begin{itemize}
    \item We propose a novel token swapping operation that fully exploits semantic correlations of nodes to generate diverse token sequences.
    \item We develop a Transformer-based backbone with a center alignment loss to learn node representations from the generated diverse token sequences.
    \item Extensive experiments on various datasets with different training data ratios showcase the effectiveness of \name in node classification. 
\end{itemize}

%% file: Body/1RW.tex
\subsection{Graph Neural Networks}
GNNs~\cite{gnn3,gnn2,rlp,pamt,ncn} have shown remarkable performance in this task. 
Previous studies \cite{gat,jknet,sgc,appnp} have primarily concentrated on the incorporation of diverse graph structural information into the message-passing framework. 
Classic deep learning techniques, such as the attention mechanism \cite{gat,gatv2} and residual connections \cite{jknet,gcnii}, have been exploited to enhance the information aggregation on graphs. 
Moreover, aggregating information from high-order neighbors \cite{appnp,mixhop,h2gnn} or nodes with high similarity across different feature spaces \cite{geomgcn} has been demonstrated to be efficacious in improving model performance.

Follow-up GNNs have focused on the utilization of complex graph features to extract distinctive node representations. 
A prevalent strategy entails the utilization of signed aggregation weights \cite{fagcn,gprgnn,acmgnn,glognn} to optimize the aggregation operation. 
In this way, positive and negative values are respectively associated with low- and high-frequency information, thereby enhancing the discriminative power of the learned node representations.
Nevertheless, restricted by the inherent limitations of message-passing mechanism, the potential of GNNs for graph data mining has been inevitably weakened.
Developing a new graph deep learning paradigm has attracted great attention in graph representation learning.

\subsection{Graph Transformers}
GTs~\cite{polynormer,agt,cob} have emerged as a novel architecture for graph representation learning and have exhibited substantial potential in node classification. 
A commonly adopted design paradigm for GTs is the combination of Transformer modules with GNN-style modules to construct hybrid neural network layers, called hybrid GTs~\cite{nodeformer,sgformer,specformer}. 
In this design, Transformer is employed to capture global information, while GNNs are utilized for local information extraction~\cite{graphgps,polynormer,signgt}.
Despite effectiveness, directly utilizing Transformer to model the interactions of all node pairs could occur the over-globalization issue~\cite{cob}, inevitably weakening the potential for graph representation learning.

An alternative yet effective design of GTs involves transforming the input graph into independent token sequences termed tokenized GTs \cite{ansgt,nagphormer,vcrgt,polyformer,ntformer}, which are then fed into the Transformer layer for node representation learning. 
Neighborhood tokens~\cite{nagphormer,nag+,polyformer,vcrgt,ntformer} and node tokens~\cite{ansgt,ntformer,vcrgt} are two typical elements in existing tokenized GTs.
The former, generally constructed by propagation approaches, such as random walk~\cite{nagphormer,nag+} and personalized PageRank~\cite{vcrgt}.
The latter is generated by diverse sampling methods based different similarity measurements, such as PageRank score~\cite{vcrgt} and attribute similarity~\cite{ansgt}. 
Since tokenized GTs only focus on the generated tokens, they naturally avoiding the over-globalization issue.

As pointed out in previous study~\cite{vcrgt}, node token oriented GTs are more efficient in capturing various graph information, such as long-range dependencies and heterophily, compared to neighborhood token oriented GTs.
However, we identify that previous methods only leverage a small subset of nodes as tokens for node representation learning, which could limit the model ability of deeply exploring graph information.
In this paper, we develop a new method \name that introduces a novel token swapping operation to produce more informative token sequences, further enhancing the model performance.

%% file: Body/2Pre.tex
\subsection{Node Classification}
Suppose an attributed graph is denoted as $\mathcal{G}=(V, E, \mathbf{X})$ where $V$ and $E$ are the sets of nodes and edges in the graph.
$\mathbf{X} \in \mathbb{R}^{n \times d}$ is the attribute feature matrix, where $n$ and $d$ are the number of nodes and the dimension of the attribute feature vector, respectively.
We also have the adjacency matrix $\mathbf{A} = \{0, 1\}^{n\times n}$.
If there is an edge between nodes $v_i$ and $v_j$, $\mathbf{A}_{ij} = 1$; otherwise, $\mathbf{A}_{ij} = 0$. 
$\hat{\mathbf{A}}$ denotes the normalized version calculated as $\hat{\mathbf{A}}=(\mathbf{D}+\mathbf{I})^{-1/2}(\mathbf{A}+\mathbf{I})(\mathbf{D}+\mathbf{I})^{-1/2}$ where $\mathbf{D}$ and $\mathbf{I}$ are the diagonal degree matrix and the identity matrix, respectively.
In the scenario of node classification, each node is associated with a one-hot vector to identify the unique label information, resulting in a label matrix $\mathbf{Y} \in \mathbb{R}^{n \times c}$ where $c$ is the number of labels.
Given a set of labeled nodes $V_L$, the goal of the task is to predict the labels of the rest  nodes in $V-V_L$.

\subsection{Transformer}
Here, we introduce the design of the Transformer layer, which is the key module in most GTs. 
There are two core components of a Transformer layer~\cite{transformer}, named multi-head self-attention (MSA) and feed-forward network (FFN).
Given the model input $\mathbf{H}^{n\times d}$, the calculation of MSA is as follows:
\begin{equation}
    \mathrm{MSA}(\mathbf{H}) = (||_{i=1}^{m}head_i)\cdot \mathbf{W}_{o},
    \label{eq:msa}
\end{equation}
\begin{equation}
    head_i = \mathrm{softmax}\left(\frac{[(\mathbf{H}\cdot\mathbf{W}^{Q}_{i}) \cdot (\mathbf{H}\cdot\mathbf{W}^{K}_{i})^{\mathrm{T}}]}{\sqrt{d_k}}\right)\cdot (\mathbf{H}\cdot\mathbf{W}^{V}_{i}), 
    \label{eq:single-head}
\end{equation}
where $\mathbf{W}^{Q}_{i}$, $\mathbf{W}^{K}_{i}$ and $\mathbf{W}^{V}_{i}$ are the learnable parameter matrices of the $i$-th attention head.
$m$ is the number of attention heads.
$||$ denotes the vector concatenation operation.
$\mathbf{W}_{o}$ denotes a projection layer to obtain the final output of MSA.

FFN is constructed by two linear layers and one non-linear activation function:
\begin{equation}
    \mathrm{FFN}(\mathbf{H}) = \sigma(\mathbf{H}\cdot\mathbf{W}^{1})\cdot\mathbf{W}^{2},
    \label{eq:ffn}
\end{equation}
where $\mathbf{W}^{1}$ and $\mathbf{W}^{2}$ denote learnable parameters of the two linear layers and $\sigma(\cdot)$ denotes the GELU activation function.

%% file: Body/3Method.tex
\begin{figure*}[t]
\centering
\includegraphics[width=16cm]{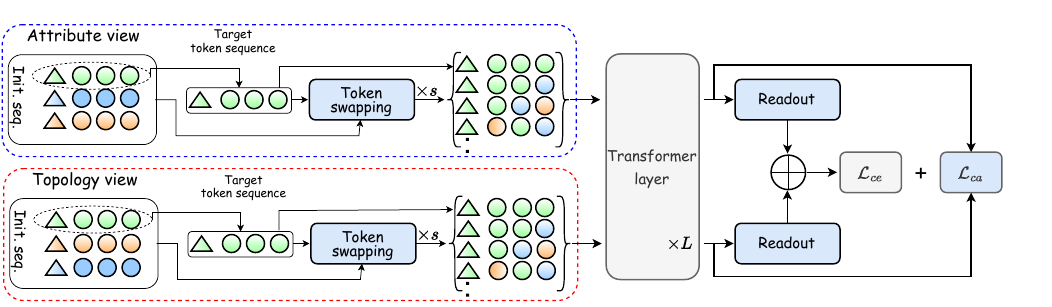}
\caption{
The overall framework of \name.
First, we generate the initial token sequences from both the attribute view and topology view. Then, we utilize the proposed token swapping operation to generate new token sequences for each target node. These generated token sequences are then fed into a Transformer-based backbone to learn node representations and generate predicted labels. Additionally, a center alignment loss is adopted to further constrain the representations extracted from different token sequences.
}
\label{fig:fw}
\end{figure*}


In this section, we detail our proposed \name.
Specifically, \name conducts the token sampling operation based on different feature spaces to generate the initial token sequences.
Then, \name develops a novel token swapping operation that produces diverse token sequences based on the initial token sequences.
Finally, \name introduces a Transformer-based backbone with a center alignment loss to learn node representations from generated token sequences.
The overall framework of \name is shown in Figure \ref{fig:fw}.

\subsection{Token Sampling}
Token sampling, which aims to select relevant nodes from input graph to construct a token sequence for each target node, is a crucial operation in recent GTs~\cite{ansgt,vcrgt}.
Generally speaking, the token sampling operation for the target node $v_i$ can be summarized as follows:
\begin{equation}
    N_i = \{v_j | v_j \in \mathrm{Top}(\mathbf{S}_i, k)\},
    \label{eq:token_sampling}
\end{equation}
where $N_i$ denotes the token set of node $v_i$.
$\mathrm{Top}(\cdot)$ denotes the top-$k$ sampling operation and $k$ is the number of sampled tokens. 
$\mathbf{S}_i \in \mathbb{R}^{1\times n}$ denotes the vector of similarity scores between $v_i$ and other nodes.
Obviously, different strategies for measuring node similarity will result in different sets of sampled tokens. 

According to empirical results in previous studies~\cite{ansgt,vcrgt}, measuring the node similarity from different feature spaces to sample different types of tokens can effectively enhance the model performance.
Hence, in this paper, we sample the tokens from both the attribute feature space and the topology feature space, resulting in two token sets $N^{A}_i$ and $N^{T}_i$, respectively.

Specifically, for $N^{A}_i$, we utilize the raw attribute features to calculate the cosine similarity of each node pair to obtain the similarity scores.
For $N^{T}_i$, we adopt the neighborhood features to represent the topology features of nodes.
The neighborhood feature matrix is calculated as $\mathbf{X}^{\prime}= \phi(\hat{\mathbf{A}}, \mathbf{X}, K)$ where $K$ denotes the number of propagation steps and $\phi(\cdot)$ denotes the personalized PageRank propagation~\cite{appnp}.
Then we leverage the neighborhood features to measure the similarity of nodes in the topology feature space via the cosine similarity.

\subsection{Token Swapping}\label{sec:swapping}
As discussed in \autoref{fig:motivation}, existing node token generators~\cite{gophormer,ansgt,vcrgt} could be regarded as selecting the 1-hop neighborhood nodes in the constructed $k$-NN graph, which is inefficient in fully leveraging the semantic relevance between nodes and restricts the diversity of the token sequences, further limiting the model performance. 

To effectively obtain diverse token sequences, \name introduces a novel operation based on the unique characteristic of token sequences, called token swapping.
The key idea is to swap tokens in different sequences to construct new token sequences.
Specifically, for each node token $v_j$ in the token set of node $v_i$, we generate a new node token $v^{new}= \zeta(N_j)$ based on the token set $N_j$ of $v_j$, where $\zeta(\cdot)$ denotes the random sampling operation.
Then, the new token set $N^{\prime}_i$ of $v_i$ is generated as follows:
\begin{equation}
    N^{\prime}_i=\{\zeta(N_j)|v_j\in N_i\}.
    \label{eq:condidate_token}
\end{equation}

\begin{figure}[t]
\centering
\includegraphics[width=7.5cm]{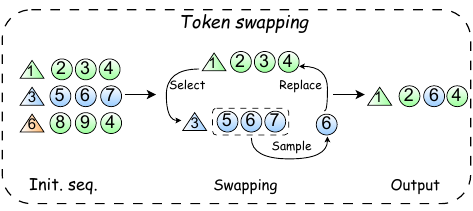}
\caption{
Illustration of the token swapping, where node 1 is the target node. We first select node 3 and regard the tokens in its token sequences as the candidates.
Then we select node 6 from the candidates to swap node 3, and construct the new token sequence.
}
\label{fig:swap}
\end{figure}

\autoref{eq:condidate_token} indicates that for each node token $v_j$, we swap it with a random node $v^{new}$ in its node token set $N_j$ to construct the new token set $N^{\prime}_i$ for the target node $v_i$.
\autoref{fig:swap} provides a toy example to illustrate the token swapping operation.

Here, we provide deep insights for the token swapping operation.
According to \autoref{fig:motivation}, nodes in the token set are the 1-hop neighbors of the target node in the $k$-NN graph.
Therefore, the selected node $v^{new}$ is within the 2-hop neighborhoods of $v_i$.
Therefore, performing the swapping operation $t$ times is equal to enlarge the sampling space from 1-hop neighborhood to $(t+1)$-hop neighborhood.
Hence, the swapping operation effectively utilize the semantic relevance between nodes to generate diverse token sequences.
The overall algorithm of token swapping is summarized in Algorithm \ref{alg:tokenswap}.

\begin{algorithm}[tb]
   \caption{The Token Swapping Algorithm}
   \label{alg:tokenswap}
\begin{algorithmic}[1]
   \REQUIRE Sampled token set of all nodes $N\in \mathbb{R}^{n\times k}$; Target node $v_i$; Probability $p$; Swapping times $t$
   \ENSURE The new token set $N^{\prime}_i\in \mathbb{R}^{1\times k}$ of $v_i$

    \STATE Initialize $N^{\prime}_i=N_i$;
    \FOR{$t_0=1$ {\bfseries to} $t$}
    \STATE Initialize $N^{new}=\{\}$;
   \FOR{$v_j \in N^{\prime}_i$ }
   \IF{$random(0,1)>p$}
   \STATE $N^{new}=N^{new}\cup\{v_j\}$;
    \ELSE
    \STATE $v^{new}= \zeta(N_j)$;
    \STATE $N^{new}=N^{new}\cup\{v^{new}\}$;

   \ENDIF
   \ENDFOR
   \STATE $N^{\prime}_i = N^{new}$;
   \ENDFOR
   \RETURN $N^{\prime}_i$

\end{algorithmic}
\end{algorithm}

After generating various token sets, we utilize them to construct the input token sequence for representation learning.
Given a token set $N_i$, the input token sequence $\mathbf{Z}_i^s$ of node $v_i$ is as:
\begin{equation}
    \mathbf{Z}_i^s = [\mathbf{X}_i, \mathbf{X}_{N_{i,0}}, \dots, \mathbf{X}_{N_{i,k-1}}].
    \label{eq:input_token}
\end{equation}
By performing Algorithm \ref{alg:tokenswap} $s$ times, we can finally obtain $1+s$ token sets for the target node $v_i$.
We combine all token sequences generated by \autoref{eq:input_token} based on different token sets to obtain the final input token sequences $\mathbf{Z}_i\in \mathbb{R}^{(1+s)\times (1+k) \times d}$.
Following the same process, the input token sequences generated by $N^{A}_i$ and $N^{T}_i$ are denoted as $\mathbf{Z}^A_i$ and $\mathbf{Z}^T_i$, respectively.

\subsection{Transformer-based Backbone}
The proposed Transformer-based backbone aims to learn node representations and predict the node labels according to the input token sequences.
Take the input sequence $\mathbf{Z}^A_i$ for example, we first utilize the projection layer to obtain the model input:
\begin{equation}
    \mathbf{Z}^{A,(0)}_i = \rho(\mathbf{Z}^A_i),
    \label{eq:projection}
\end{equation}
where $\rho(\cdot)$ denotes the projection layer and $\mathbf{Z}^{A,(0)}_i \in \mathbb{R}^{(1+s)\times k \times d_0}$ denotes the model input of node $v_i$.

Then, a Transformer layer-based encoder is applied to learn node representations from the model input:
\begin{align}
\mathbf{Z}^{\prime A,(l)}_{i} &=\operatorname{MSA}\left(\operatorname{LN}\left(\mathbf{Z}^{A,(l-1)}_{i}\right)\right)+\mathbf{Z}^{A, (l-1)}_{i}, \\
\mathbf{Z}^{A,(l)}_{i} &=\operatorname{FFN}\left(\operatorname{LN}\left(\mathbf{Z}^{\prime A,(l)}_{i}\right)\right)+\mathbf{Z}^{\prime A,(l)}_{i}, 
\end{align}
where $l=1, \ldots, L$ indicates the $l$-th Transformer layer and $\mathrm{LN}(\cdot)$ denotes the LayerNorm operation.

Through the encoder, we obtain the representations of input token sequences $\mathbf{Z}^{A,(L)}_i \in \mathbb{R}^{(1+s)\times (1+k) \times d_L}$. 
Then, we take the representation of the first item in each token sequence as the final representation of the input sequence.
This is because the first item in each token sequence is the target node itself, and the output representation has learned necessary information from other sampling tokens in the input sequence via the Transformer encoder.

Hence, the final output of the Transformer encoder is denoted as $\mathbf{Z}^{A,i} \in \mathbb{R}^{(1+s) \times d_L}$.
Then, we utilize the following readout function to obtain the node representation learned from multi token sequences:
\begin{equation}
    \mathbf{Z}^{A}_i = \mathbf{Z}^{A,i}_0||( \frac{1}{s}\sum_{j=1}^{s} \mathbf{Z}^{A,i}_j),
    \label{eq:readout}
\end{equation}
where $||$ denotes the concatenation operation, $\mathbf{Z}^{A}_i\in \mathbb{R}^{1\times d_L}$ denotes the representation of $v_i$ learned from token sequences generated from the attribute feature space.
Similarly, we can obtain $\mathbf{Z}^{T}_i$ from the topology feature space.

To effectively utilize information of different feature spaces, we leverage the following strategy to fuse the learned representations:
\begin{equation}
    \mathbf{Z}^{F}_i = \alpha\cdot\mathbf{Z}^{A}_i + (1-\alpha)\cdot\mathbf{Z}^{T}_i,
    \label{eq:finalrepre}
\end{equation}
where $\alpha \in [0, 1]$ is a hyper-parameter to balance the contribution from attribute features and topology features on the final node representation.

At the end, we adopt Multi-Layer Perception-based predictor for label prediction and the cross-entropy loss for model training:
\begin{equation}
    \mathbf{Y}^{\prime}_i = \mathrm{MLP}(\mathbf{Z}^{F}_i),
    \label{eq:prediction}
\end{equation}
\begin{equation}
        \mathcal{L}_{ce} = -\sum_{i\in V_{L}}\sum_{j = 0}^{c} {\mathbf{Y}_{i,j}}\mathrm{ln}\mathbf{Y}^{\prime}_{i,j}.
    \label{eq:celoss}
\end{equation}

\input{Tab/nc_dense}

\subsection{Center Alignment Loss}
To further enhance the model's generalization, we develop a center alignment loss to constrain the representations learned from different token sequences for each node.
Specifically, given the representations of multi-token sequences $\mathbf{Z}^{i}\in \mathbb{R}^{(1+s) \times d_L}$, we first calculate the center representation $\mathbf{Z}^{i}_c = \frac{1}{(1+s)}\sum_{j=0}^{s} \mathbf{Z}^{i}_j$.
Then, the center alignment loss is calculated as follows:
\begin{equation}
        \mathrm{CAL}(\mathbf{Z}^{i}, \mathbf{Z}^{i}_c) = 1 - \frac{1}{(1+s)}\sum_{j=0}^{s} \mathrm{Cosine}(\mathbf{Z}^{i}_j, \mathbf{Z}^{i}_c),
    \label{eq:cal}
\end{equation}
where $\mathrm{Cosine}(\cdot)$ denotes the cosine similarity function.

The rationale of \autoref{eq:cal} is that the representations learned from different token sequences can be regarded as different views of the target node.
Therefore, these representations should naturally be close to each other in the latent space.
In practice, we separately calculate the center alignment loss of token sequences from different feature spaces:
\begin{equation}
        \mathcal{L}_{ca} = \mathrm{CAL}(\mathbf{Z}^{A,i}, \mathbf{Z}^{A,i}_c) + \mathrm{CAL}(\mathbf{Z}^{T,i}, \mathbf{Z}^{T,i}_c).
    \label{eq:calloss}
\end{equation}
The overall loss of \name is as follows:
\begin{equation}
        \mathcal{L} = \mathcal{L}_{ce} + \lambda\cdot\mathcal{L}_{ca},
    \label{eq:allloss}
\end{equation}
where $\lambda$ is a coefficient controlling the balance between the two loss functions.

%% file: Tab/nc_dense.tex
\begin{table*}[ht]
\centering
\caption{Comparison of all models in terms of mean accuracy $\pm$ stdev (\%) under dense splitting. The best results appear in \textbf{bold}. The second results appear in \underline{underline}.}
\scalebox{0.80}{
\begin{tabular}{lcccccccccccc}
\toprule
Dataset& Photo & ACM & Computer  &Citeseer &WikiCS& BlogCatalog & UAI2010 & Flickr   \\
$\mathcal{H}$& 0.83 & 0.82 & 0.78  &0.74 & 0.66& 0.40 & 0.36& 0.24  \\ \hline

SGC& 93.74\tiny{$\pm$0.07} & 93.24\tiny{$\pm$0.49} &88.90\tiny{$\pm$0.11} & 76.81\tiny{$\pm$0.26} & 76.67\tiny{$\pm$0.19}& 72.61\tiny{$\pm$0.07} &69.87\tiny{$\pm$0.17} &47.48\tiny{$\pm$0.40}  \\

APPNP&{94.98\tiny{$\pm$0.41}}& 93.00\tiny{$\pm$0.55} &\underline{91.31\tiny{$\pm$0.29} } & {77.52\tiny{$\pm$0.22}}  & 81.96\tiny{$\pm$0.14}
& 94.77\tiny{$\pm$0.19} &{77.41\tiny{$\pm$0.47}} & 84.66\tiny{$\pm$0.31}  \\

GPRGNN& 94.57\tiny{$\pm$0.44} & 93.42\tiny{$\pm$0.20} &90.15\tiny{$\pm$0.34}  & 77.59\tiny{$\pm$0.36} &82.43\tiny{$\pm$0.29}& {94.36\tiny{$\pm$0.29} }&76.94\tiny{$\pm$0.64} &{85.91\tiny{$\pm$0.51}}    \\

FAGCN& 94.06\tiny{$\pm$0.03} & 93.37\tiny{$\pm$0.24} &83.17\tiny{$\pm$1.81} & 76.19\tiny{$\pm$0.62} 
& 79.89\tiny{$\pm$0.93}& 79.92\tiny{$\pm$4.39} &72.17\tiny{$\pm$1.57} & 82.03\tiny{$\pm$0.40}   \\

BM-GCN& 95.10\tiny{$\pm$0.20} &{93.68\tiny{$\pm$0.34} }&91.28\tiny{$\pm$0.96} & 77.91\tiny{$\pm$0.58} & {83.90\tiny{$\pm$0.41} }&94.85\tiny{$\pm$0.42} & 77.39\tiny{$\pm$1.13} &   83.97\tiny{$\pm$0.87} \\

ACM-GCN& 94.56\tiny{$\pm$0.21} & 93.04\tiny{$\pm$1.28} &85.19\tiny{$\pm$2.26} &77.62\tiny{$\pm$0.81} 
&\underline{83.95\tiny{$\pm$0.41}}
& 94.53\tiny{$\pm$0.53} &76.87\tiny{$\pm$1.42} & 83.85\tiny{$\pm$0.73}  \\

\hline
NAGphormer&  \underline{95.47\tiny{$\pm$0.29}} & 93.32\tiny{$\pm$0.30} &90.79\tiny{$\pm$0.45} &  77.68\tiny{$\pm$0.73}& 
83.61\tiny{$\pm$0.28} & 94.42\tiny{$\pm$0.63} &76.36\tiny{$\pm$1.12} & 86.85\tiny{$\pm$0.85}  \\

SGFormer& 92.93\tiny{$\pm$0.12} & 93.79\tiny{$\pm$0.34} &81.86\tiny{$\pm$3.82} &  77.86\tiny{$\pm$0.76}& 79.65\tiny{$\pm$0.31} & 94.33\tiny{$\pm$0.19} &57.98\tiny{$\pm$3.95} & 61.05\tiny{$\pm$0.68}  \\

Specformer& 95.22\tiny{$\pm$0.13} & 93.63\tiny{$\pm$1.94} &85.47\tiny{$\pm$1.44} & 77.96\tiny{$\pm$0.89}&  83.74\tiny{$\pm$0.62}  & 94.21\tiny{$\pm$0.23} &73.06\tiny{$\pm$0.77} & 86.55\tiny{$\pm$0.40} \\

VCR-Graphormer
&95.38\tiny{$\pm$0.51} & 93.11\tiny{$\pm$0.79} &90.47\tiny{$\pm$0.58} & 77.21\tiny{$\pm$0.65}& 80.82\tiny{$\pm$0.72} & 94.19\tiny{$\pm$0.17} &76.08\tiny{$\pm$0.52} & 85.96\tiny{$\pm$0.55}    \\

PolyFormer
& 95.45\tiny{$\pm$0.21} & \underline{94.27\tiny{$\pm$0.44}} &90.87\tiny{$\pm$0.74} & \underline{78.03\tiny{$\pm$0.86}}& 83.79\tiny{$\pm$0.75} & \underline{95.08\tiny{$\pm$0.43} }&\underline{77.92\tiny{$\pm$0.82} }& \underline{87.01\tiny{$\pm$0.57} }   \\

\hline

\name & 
\textbf{95.92\tiny{$\pm$0.18}} & \textbf{94.98\tiny{$\pm$0.41}} & 
\textbf{91.73\tiny{$\pm$0.72}} & 
\textbf{78.49\tiny{$\pm$0.95}} & 
\textbf{84.52\tiny{$\pm$0.63}}& \textbf{95.93\tiny{$\pm$0.56}} & \textbf{79.06\tiny{$\pm$0.73}} & \textbf{87.56\tiny{$\pm$0.61}}  \\     
 \toprule
\end{tabular}
}

\label{tab:dense-ncre}
\end{table*}

%% file: Body/4Exp.tex


\subsection{Dataset}
We adopt eight widely used datasets, involving homophily and heterophily graphs: 
Photo~\cite{nagphormer}, ACM~\cite{acm}, Computer~\cite{nagphormer}, BlogCatalog~\cite{socialnets}, UAI2010~\cite{amgcn}, Flickr~\cite{socialnets} and Wiki-CS~\cite{roman}.
The edge homophily ratio~\cite{glognn} ${H}(\mathcal{G})\in[0,1]$ is adopted to evaluate the graph's homophily level. 
${H}(\mathcal{G}) \rightarrow 1$ means strong homophily, 
while ${H}(\mathcal{G}) \rightarrow 0$ means strong heterophily.
Statistics of datasets are summarized in Appendix \ref{app:data}.
To comprehensively evaluate the model performance in node classification, we provide two strategies to split datasets, called dense splitting and sparse splitting.
In dense splitting, we randomly choose 50\% of each label as the training set, 25\% as the validation set, and the rest as the test set, which is a common setting is previous studies~\cite{nodeformer,sgformer}.
While in sparse splitting~\cite{gprgnn}, we adopt 2.5\%/2.5\%/95\% splitting for training set, validation set and test set, respectively.

\subsection{Baseline}
We adopt eleven representative approaches as the baselines: SGC~\cite{sgc}, APPNP~\cite{appnp}, GPRGNN~\cite{gprgnn}, FAGCN~\cite{fagcn}, BM-GCN~\cite{bmgcn}, ACM-GCN~\cite{acmgnn}, NAGphormer~\cite{nagphormer}, SGFormer~\cite{sgformer}, Specformer~\cite{specformer}, VCR-Graphormer~\cite{vcrgt} and PolyFormer~\cite{polyformer}.
The first six are mainstream GNNs and others are representative GTs.

\input{Tab/nc_sparse}

\subsection{Performance Comparison}
To evaluate the model performance in node classification, we run each model ten times with random initializations. The results in terms of mean accuracy and standard deviation are reported in \autoref{tab:dense-ncre} and \autoref{tab:sparse-ncre}.

First, we can observe that \name achieves the best performance on all datasets with different data splitting strategies, demonstrating the effectiveness of \name in node classification.
Then, we can find that advanced GTs obtain more competitive performance than GNNs on over half datasets under dense splitting.
But under sparse splitting, the situation reversed.
An intuitive explanation is that Transformer has more learnable parameters than GNNs, which bring more powerful modeling capacity.
However, it also requires more training data than GNNs in the training stage to ensure the performance.
Therefore, when the training data is sufficient, GTs can achieve promising performance.
And when the training data is sparse, GTs usually leg behind GNNs.
Our proposed \name addresses this issue by introducing the token swapping operation to generate diverse token sequences. 
This operation effectively augments the training data, ensuring the model training even in the sparse data scenario.
In addition, the tailored center alignment loss also constrains the model parameter learning, further enhancing the model performance.

\begin{figure}[t]
\centering
\includegraphics[width=7.5cm]{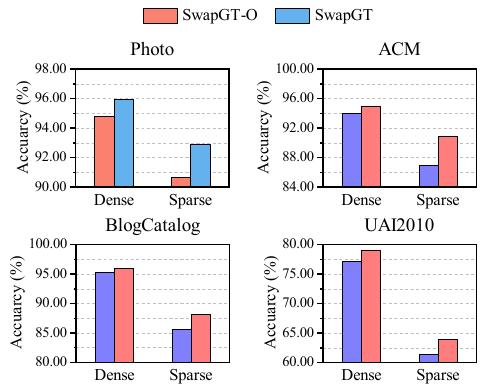}
\caption{
Performances of \name with or without the center alignment loss.
}
\label{fig:align}
\end{figure}

\begin{figure}[t]
\centering
\includegraphics[width=7.5cm]{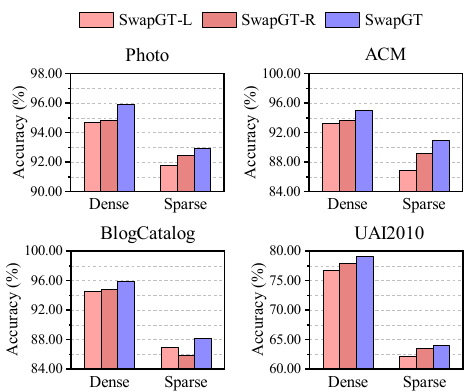}
\caption{
Performances of \name with different token sequence generation strategies.}
\label{fig:ts}
\end{figure}

\begin{figure}[t]
\centering
\includegraphics[width=7.3cm]{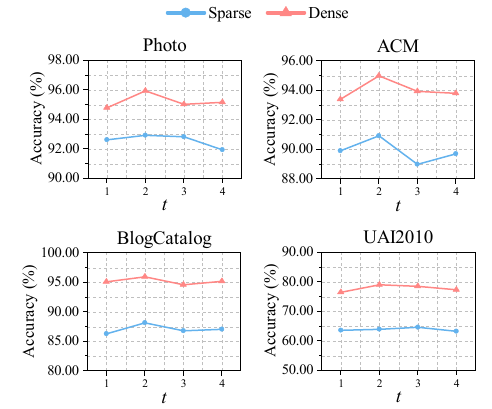}
\caption{
Analysis on the swapping times $t$.}
\label{fig:t}
\end{figure}

\begin{figure}[t]
\centering
\includegraphics[width=7.3cm]{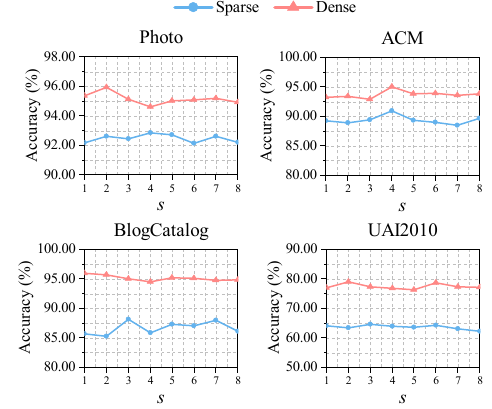}
\caption{
Analysis on the augmentation times $s$.}
\label{fig:s}
\end{figure}

\subsection{Study on the center alignment loss}
The center alignment loss, proposed to constrain the representation learning from multiple token sequences, is a key design of \name.
Here, we validate the effectiveness of the center alignment loss in node classification.
Specifically, we develop a variant of \name by removing the center alignment loss, called \name-O.
Then, we evaluate the performance of \name-O on all datasets under dense splitting and sparse splitting.
Due to the space limitation, we only report the results on four datasets in \autoref{fig:align}, other results are reported in Appendix \ref{app:exp-ca}.
"Den." and "Spa." denotes the experimental results under dense splitting and sparse splitting, respectively.
Based on the experimental results, we can have the following observations:
1) \name beats \name-O on all datasets, indicating that the developed center alignment loss can effectively enhance the performance of \name.
2) Adopting the center alignment loss can bring more significant improvements in sparse setting than those in dense setting.
This situation implies that introducing the reasonable constraint loss function based on the property of node token sequences can effectively improve the model training when the training data is sparse.

\subsection{Study on the token sequence generation}\label{exp:ts}
The generation of token sequences is another key module of \name, which develops a novel token swapping operation can fully leverage the semantic relevance of nodes to generate informative token sequences.
In this section, we evaluate the effectiveness of the proposed strategy by comparing it with two naive strategies.
One is to enlarge the sampling size $k$. We propose a variant called \name-L by sampling $2k$ tokens to construct token sequences.
The other is to randomly sample $k$ tokens from the enlarged $2k$ token set to construct multiple token sequences, called \name-R.
Performance of these variants are shown in \autoref{fig:ts} and results on other datasets are reported in Appendix \ref{app:exp-ts}.

We can observe that \name-R outperforms \name-L on most cases, indicating that constructing multiple token sequences is better for node representation learning of tokenized GTs than generating single long token sequence. 
Moreover, \name surpasses \name-R on all cases, showcasing the superiority of the proposed token swapping operation in generation of multiple token sequences.
This observation also implies that constructing informative token sequences can effectively improve the performance of tokenized GTs.

\subsection{Analysis on the swapping times $t$}
As discussed in Section \ref{sec:swapping}, $t$ determines the range of candidate tokens from the constructed $k$-NN graph, further affecting the model performance.
To validate the influence of $t$ on model performance, we vary $t$ in $\{1,2,3,4\}$ and observe the changes of model performance.
Results are shown in \autoref{fig:t} and Appendix \ref{app:exp-t}.
We can clearly observe that \name can achieve satisfied performance on all datasets when $t$ is no less than 2.
This situation indicates that learning from tokens with semantic associations beyond the immediate neighbors can effectively enhancing the model performance.
This phenomenon also reveals that reasonably enlarging  the sampling space to seek more informative tokens is a promising way to improve the effect of node tokenized GTs.

\subsection{Analysis on the augmentation times $s$}
The augmentation times $s$ determines how many token sequences are adopted for node representation learning. Similar to $t$, we vary $s$ in $\{1,2,\dots,8\}$ and report the performance of \name. 
Results are shown in \autoref{fig:s} and Appendix \ref{app:exp-s}.
Generally speaking, sparse splitting requires a larger $s$ to achieve the best performance, compared to dense splitting.
This is because \name needs more token sequences for model training in the sparse data scenario.
This situation indicates that a tailored data augmentation strategy can effectively improve the performance of tokenized GTs when training data is sparse.
Moreover, the optimal $s$ varies on different graphs. 
This is because different graphs exhibit different topology features and attribute features, which affects the generation of token sequences, further influencing the model performance.

%% file: Tab/nc_sparse.tex
\begin{table*}[ht]
\centering
\caption{Comparison of all models in terms of mean accuracy $\pm$ stdev (\%) under sparse splitting. The best results appear in \textbf{bold}. The second results appear in \underline{underline}.}
\scalebox{0.8}{
\begin{tabular}{lcccccccccccc}
\toprule
Dataset& Photo & ACM & Computer  &Citeseer &WikiCS& BlogCatalog & UAI2010 & Flickr   \\
$\mathcal{H}$& 0.83 & 0.82 & 0.78  &0.74 & 0.66& 0.40 & 0.36& 0.24  \\ \hline

SGC& 91.90\tiny{$\pm$0.35} & 89.57\tiny{$\pm$0.28} &86.79\tiny{$\pm$0.19} & 66.41\tiny{$\pm$0.59} & 74.99\tiny{$\pm$0.19}& 71.23\tiny{$\pm$0.06} &51.61\tiny{$\pm$0.41} &39.43\tiny{$\pm$0.50}  \\

APPNP& \underline{92.24\tiny{$\pm$0.28}}& 89.91\tiny{$\pm$0.89} &\underline{87.64\tiny{$\pm$0.39} } & \underline{66.70\tiny{$\pm$0.11}}  & 77.42\tiny{$\pm$0.31}
& 81.76\tiny{$\pm$0.38} &\underline{61.65\tiny{$\pm$0.71}} & 71.39\tiny{$\pm$0.62}  \\

GPRGNN& 92.13\tiny{$\pm$0.32} & 89.47\tiny{$\pm$0.90} &86.38\tiny{$\pm$0.44}  & 66.50\tiny{$\pm$0.62} & 77.59\tiny{$\pm$0.49}& \underline{84.57\tiny{$\pm$0.35} }&58.75\tiny{$\pm$0.75} &\underline{71.89\tiny{$\pm$0.89}}    \\

FAGCN& 92.02\tiny{$\pm$0.18} & 88.47\tiny{$\pm$0.31} &83.99\tiny{$\pm$1.95} & 64.54\tiny{$\pm$0.66}  & 75.21\tiny{$\pm$0.84}& 76.38\tiny{$\pm$0.82} &54.67\tiny{$\pm$0.96} & 63.68\tiny{$\pm$0.72}  \\

BM-GCN& 91.19\tiny{$\pm$0.39} &\underline{90.11\tiny{$\pm$0.60} }&86.14\tiny{$\pm$0.51} & 66.11\tiny{$\pm$0.47}& {77.39\tiny{$\pm$0.37} } &84.05\tiny{$\pm$0.54} & 57.51\tiny{$\pm$1.14} &   60.82\tiny{$\pm$0.76} \\

ACM-GCN& 91.71\tiny{$\pm$0.64} & 89.68\tiny{$\pm$0.45} &86.64\tiny{$\pm$0.59} &64.85\tiny{$\pm$1.19}& \underline{77.68\tiny{$\pm$0.57}} & 77.17\tiny{$\pm$1.34} &56.05\tiny{$\pm$2.11} & 64.58\tiny{$\pm$1.53}  \\

\hline
NAGphormer& 91.65\tiny{$\pm$0.80} & 89.73\tiny{$\pm$0.48} &85.31\tiny{$\pm$0.65} &  63.66\tiny{$\pm$1.68}& 76.93\tiny{$\pm$0.75} & 79.19\tiny{$\pm$0.41} &58.36\tiny{$\pm$1.01} & 67.48\tiny{$\pm$1.04}  \\

SGFormer& 90.13\tiny{$\pm$0.56} & 88.03\tiny{$\pm$0.60} &80.07\tiny{$\pm$0.21} &  62.41\tiny{$\pm$0.94}& 74.69\tiny{$\pm$0.52} & 78.15\tiny{$\pm$0.69} &50.19\tiny{$\pm$1.72} & 51.01\tiny{$\pm$1.05}  \\

Specformer& 90.57\tiny{$\pm$0.55} & 88.20\tiny{$\pm$1.05} &85.55\tiny{$\pm$0.63} & 62.64\tiny{$\pm$1.54} &  75.24\tiny{$\pm$0.71}& 79.75\tiny{$\pm$1.29} &57.42\tiny{$\pm$1.06} & 56.94\tiny{$\pm$1.48}  \\

VCR-Graphormer
& 91.39\tiny{$\pm$0.75} & 86.81\tiny{$\pm$0.84} &85.06\tiny{$\pm$0.64} & 57.61\tiny{$\pm$0.60} & 72.81\tiny{$\pm$1.44} & 74.90\tiny{$\pm$1.18} &56.43\tiny{$\pm$1.10} & 50.93\tiny{$\pm$1.12}   \\

PolyFormer
& 91.52\tiny{$\pm$0.78} & 89.83\tiny{$\pm$0.62} &85.75\tiny{$\pm$0.78} & 64.77\tiny{$\pm$1.27} & 75.12\tiny{$\pm$1.16}& 81.02\tiny{$\pm$0.81} &58.89\tiny{$\pm$0.77} & 67.85\tiny{$\pm$1.43}    \\

\hline

\name & 
\textbf{92.93\tiny{$\pm$0.26}} & \textbf{90.92\tiny{$\pm$0.69}} & 
\textbf{88.14\tiny{$\pm$0.52}} & 
\textbf{69.91\tiny{$\pm$1.02}} & 
\textbf{78.11\tiny{$\pm$0.83}}& \textbf{88.11\tiny{$\pm$0.58}} & \textbf{63.96\tiny{$\pm$1.09}} & \textbf{72.16\tiny{$\pm$1.19}}  \\     
 \toprule
\end{tabular}
}

\label{tab:sparse-ncre}
\end{table*}

%% file: Body/5APP.tex
\section{Experimental Settings}

\subsection{Dataset}\label{app:data}
Here we introduce datasets adopted for experiments. The detailed statistics of all datasets are reported in \autoref{tab:dataset}.
\begin{itemize}
    \item \textbf{Academic graphs}: This type of graph is formed by academic papers or authors and the citation relationships among them. Nodes in the graph represent academic papers or authors, and edges represent the citation relationships between papers or co-author relationships between two authors. The features of nodes are composed of bag-of-words vectors, which are extracted and generated from the abstracts and introductions of the academic papers. The labels of nodes correspond to the research fields of the academic papers or authors. ACM, Citeseer, WikiCS and UAI2010 belong to this type.
    \item \textbf{Co-purchase graphs}: This type of graph is constructed based on users' shopping behaviors. Nodes in the graph represent products. The edges between nodes indicate that two products are often purchased together. The features of nodes are composed of bag-of-words vectors extracted from product reviews. The category of a node corresponds to the type of goods the product belongs to. Computer and Photo belong to this type.

    \item \textbf{Social graphs}: This type of graph is formed by the activity records of users on social platforms. Nodes in the graph represent users on the social platform. The edges between nodes indicate the social relations between two users. Node features represent the text information extracted from the authors' homepage. The label of a node refers to the interest groups of users. BlogCatalog and Flickr belong to this type.
    
\end{itemize}

\input{Tab/dataset}
\subsection{Implementation Details}\label{app:imple}
For baselines, we refer to their official implementations and conduct a systematic tuning process on each dataset.
For \name, we employ a grid search strategy to identify the optimal parameter settings.
Specifically, We try the learning rate in $\{0.001, 0.005, 0.01\}$, dropout in $\{0.3, 0.5, 0.7\}$, dimension of hidden representations in $\{256, 512\}$,
$k$ in $\{4, 6, 8\}$, $\alpha$ in $\{0.1, \dots, 0.9\}$.
All experiments are implemented using Python 3.8, PyTorch 1.11, and CUDA 11.0 and executed on a Linux server with an Intel Xeon Silver 4210 processor, 256 GB of RAM, and a 2080TI GPU.

\section{Additional Experimental Results}\label{app:exp-results}
In this section, we provide the additional experimental results of ablation studies and parameter studies.

\subsection{Study of the center alignment loss}\label{app:exp-ca}
\begin{figure}[ht]
\centering
\includegraphics[width=17cm]{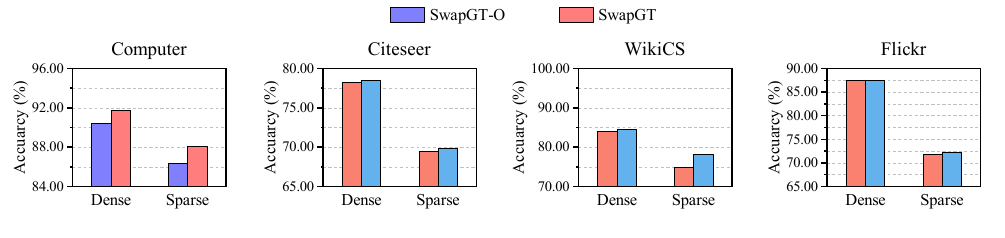}
\caption{
Performances of \name with or without the center alignment loss.
}
\label{fig:align-APP}
\end{figure}

The experimental results of \name and \name-O on the rest datasets are shown in \autoref{fig:align-APP}.
We can observe that \name outperforms \name-O on most datasets.
Moreover, the effect of applying the center alignment loss on \name in sparse splitting is more significant than that in dense splitting.
The above observations are in line with those reported in the main text.
Therefore, we can conclude that the center alignment loss can effectively enhance the performance of \name in node classification.

\subsection{Study of the token sequence generation}\label{app:exp-ts}
\begin{figure}[ht]
\centering
\includegraphics[width=17cm]{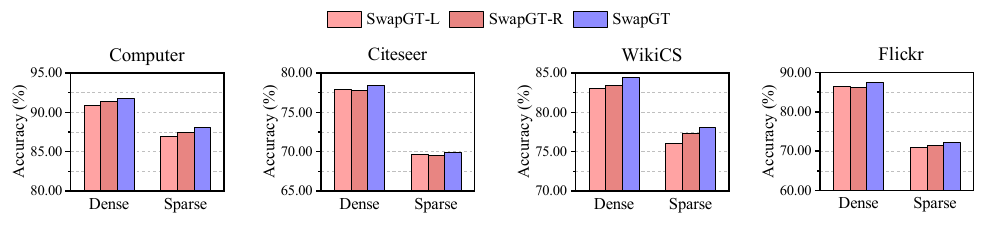}
\caption{
Performances of \name with different token sequence generation strategies.
}
\label{fig:ts-APP}
\end{figure}

The experimental results of \name with different token sequence generation strategies on the rest datasets are shown in \autoref{fig:ts-APP}.
We can find that the additional experimental results exhibit similar observations shown in the main text.
This situation demonstrates the effectiveness of the token sequence generation with the proposed token swapping operation in enhancing the performance of tokenized GTs.
Moreover, we can also observe that the gains of introducing the token swapping operation vary on different graphs based on the results shown in \autoref{fig:ts} and \autoref{fig:ts-APP}.
This phenomenon may attribute to that different graphs possess unique topology and attribute information, which further impact the selection of node tokens. 
While \name applies the uniform strategy for selecting node tokens, which could lead to varying gains of token swapping.
This situation also motivates us to consider different strategies of token selection on different graphs as the future work.

\subsection{Analysis of the swapping times $t$}\label{app:exp-t}
\begin{figure}[ht]
\centering
\includegraphics[width=17cm]{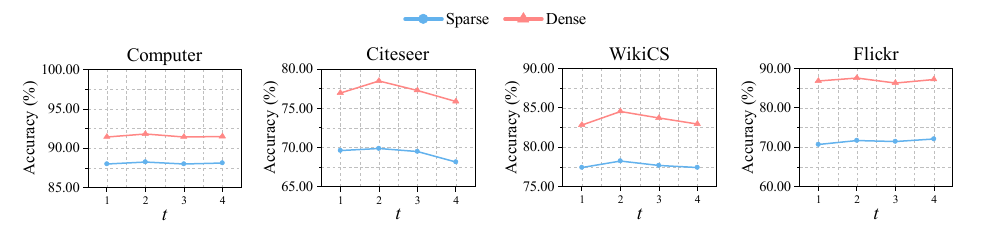}
\caption{
Performances of \name with varying $t$.
}
\label{fig:t-APP}
\end{figure}

Here we report the rest results of \name with varying $t$, which are shown in \autoref{fig:t-APP}.
Similar to the phenomenons shown in \autoref{fig:t}, \name can achieve the best performance on all datasets when $t>2$.
Based on the results shown in \autoref{fig:t-APP} and \autoref{fig:t}, we can conclude that introducing tokens beyond first-order neighbors via the proposed token swapping operation can effective improve the performance of \name in node classification.

\subsection{Analysis of the augmentation times $s$}\label{app:exp-s}
\begin{figure}[t]
\centering
\includegraphics[width=17cm]{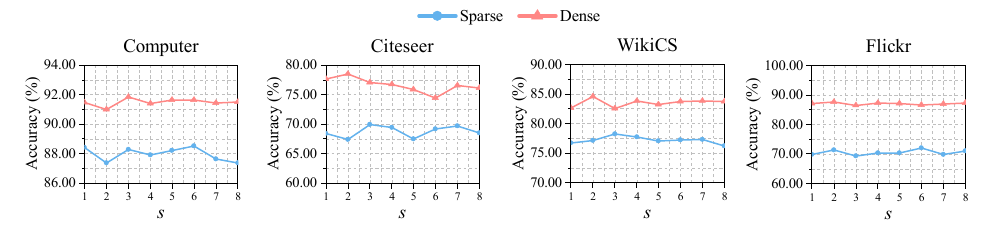}
\caption{
Performances of \name with varying $s$.
}
\label{fig:s-APP}
\end{figure}

Similar to analysis of $t$, the rest results of \name with varying $s$ are shown in \autoref{fig:s-APP}.
We can also observe the similar situations shown in \autoref{fig:s} that \name requires a larger value of $s$ under sparse splitting compared to dense splitting.
The situation demonstrates that introducing augmented token sequences can bring more significant performance gain in sparse splitting than that in dense splitting.

%% file: Tab/dataset.tex
\begin{table}[ht]
\caption{Statistics of datasets, ranked by the homophily level.}
\centering
\renewcommand\arraystretch{1.0}
\scalebox{1.}{
\begin{tabular}{lrrrrlc}
\toprule
\multicolumn{1}{l}{Dataset} & \multicolumn{1}{l}{\# nodes} & \multicolumn{1}{l}{\# edges} & \multicolumn{1}{l}{\# features} & \multicolumn{1}{l}{\# labels}  & \multicolumn{1}{c}{$\mathcal{H}\downarrow$} \\ \hline
Photo& 7,650& 238,163& 745 & 8   &      0.83     \\
ACM& 3,025& 1,3128& 1,870& 3 &   0.82  \\
Computer& 13,752& 491,722& 767& 10  &       0.78  \\
Citeseer& 3,327& 4,552& 3,703& 6  &       0.74   \\
WikiCS& 11,701& 216,123& 300 & 10      & 0.66  \\
BlogCatalog & 5,196& 171,743& 8,189 & 6       & 0.40  \\
UAI2010& 3,067 & 28,311& 4,973& 19 &       0.36   \\
Flickr& 7,575 & 239,738 & 12,047& 9       & 0.24   \\
 \toprule
\end{tabular}
}
\label{tab:dataset}
\end{table}

%% file: main.bbl
\begin{thebibliography}{42}
\providecommand{\natexlab}[1]{#1}
\providecommand{\url}[1]{\texttt{#1}}
\expandafter\ifx\csname urlstyle\endcsname\relax
  \providecommand{\doi}[1]{doi: #1}\else
  \providecommand{\doi}{doi: \begingroup \urlstyle{rm}\Url}\fi

\bibitem[Abu{-}El{-}Haija et~al.(2019)Abu{-}El{-}Haija, Perozzi, Kapoor, Alipourfard, Lerman, Harutyunyan, Steeg, and Galstyan]{mixhop}
Abu{-}El{-}Haija, S., Perozzi, B., Kapoor, A., Alipourfard, N., Lerman, K., Harutyunyan, H., Steeg, G.~V., and Galstyan, A.
\newblock Mixhop: Higher-order graph convolutional architectures via sparsified neighborhood mixing.
\newblock In \emph{Proceedings of the International Conference on Machine Learning}, 2019.

\bibitem[Bo et~al.(2021)Bo, Wang, Shi, and Shen]{fagcn}
Bo, D., Wang, X., Shi, C., and Shen, H.
\newblock Beyond low-frequency information in graph convolutional networks.
\newblock In \emph{Proceedings of the {AAAI} Conference on Artificial Intelligence}, 2021.

\bibitem[Bo et~al.(2023)Bo, Shi, Wang, and Liao]{specformer}
Bo, D., Shi, C., Wang, L., and Liao, R.
\newblock Specformer: Spectral graph neural networks meet transformers.
\newblock In \emph{Proceedings of the International Conference on Learning Representations}, 2023.

\bibitem[Brody et~al.(2022)Brody, Alon, and Yahav]{gatv2}
Brody, S., Alon, U., and Yahav, E.
\newblock How attentive are graph attention networks?
\newblock In \emph{Proceedings of the International Conference on Learning Representations}, 2022.

\bibitem[Chen et~al.(2020{\natexlab{a}})Chen, Lin, Li, Li, Zhou, and Sun]{oversm}
Chen, D., Lin, Y., Li, W., Li, P., Zhou, J., and Sun, X.
\newblock Measuring and relieving the over-smoothing problem for graph neural networks from the topological view.
\newblock In \emph{Proceedings of the {AAAI} Conference on Artificial Intelligence}, 2020{\natexlab{a}}.

\bibitem[Chen et~al.(2023{\natexlab{a}})Chen, Gao, Li, and He]{nagphormer}
Chen, J., Gao, K., Li, G., and He, K.
\newblock Nagphormer: A tokenized graph transformer for node classification in large graphs.
\newblock In \emph{Proceedings of the International Conference on Learning Representations}, 2023{\natexlab{a}}.

\bibitem[Chen et~al.(2023{\natexlab{b}})Chen, Li, Hopcroft, and He]{signgt}
Chen, J., Li, G., Hopcroft, J.~E., and He, K.
\newblock Signgt: Signed attention-based graph transformer for graph representation learning.
\newblock \emph{CoRR}, abs/2310.11025, 2023{\natexlab{b}}.

\bibitem[Chen et~al.(2024{\natexlab{a}})Chen, Jiang, and He]{ntformer}
Chen, J., Jiang, S., and He, K.
\newblock Ntformer: {A} composite node tokenized graph transformer for node classification.
\newblock \emph{CoRR}, abs/2406.19249, 2024{\natexlab{a}}.

\bibitem[Chen et~al.(2024{\natexlab{b}})Chen, Li, and He]{ncn}
Chen, J., Li, B., and He, K.
\newblock Neighborhood convolutional graph neural network.
\newblock \emph{Knowledge-Based Systems}, pp.\  111861, 2024{\natexlab{b}}.

\bibitem[Chen et~al.(2024{\natexlab{c}})Chen, Li, He, and He]{pamt}
Chen, J., Li, B., He, Q., and He, K.
\newblock Pamt: A novel propagation-based approach via adaptive similarity mask for node classification.
\newblock \emph{IEEE Transactions on Computational Social Systems}, 2024{\natexlab{c}}.

\bibitem[Chen et~al.(2024{\natexlab{d}})Chen, Liu, Gao, Li, and He]{nag+}
Chen, J., Liu, C., Gao, K., Li, G., and He, K.
\newblock Nagphormer+: A tokenized graph transformer with neighborhood augmentation for node classification in large graphs.
\newblock \emph{IEEE Transactions on Big Data}, 2024{\natexlab{d}}.

\bibitem[Chen et~al.(2020{\natexlab{b}})Chen, Wei, Huang, Ding, and Li]{gcnii}
Chen, M., Wei, Z., Huang, Z., Ding, B., and Li, Y.
\newblock Simple and deep graph convolutional networks.
\newblock In \emph{Proceedings of the International Conference on Machine Learning}, 2020{\natexlab{b}}.

\bibitem[Chien et~al.(2021)Chien, Peng, Li, and Milenkovic]{gprgnn}
Chien, E., Peng, J., Li, P., and Milenkovic, O.
\newblock {Adaptive Universal Generalized PageRank Graph Neural Network}.
\newblock In \emph{Proceedings of the International Conference on Learning Representations}, 2021.

\bibitem[Deng et~al.(2024)Deng, Yue, and Zhang]{polynormer}
Deng, C., Yue, Z., and Zhang, Z.
\newblock Polynormer: Polynomial-expressive graph transformer in linear time.
\newblock In \emph{Proceedings of the International Conference on Learning Representations}, 2024.

\bibitem[Fu et~al.(2024)Fu, Hua, Xie, Fang, Zhang, Sancak, Wu, Malevich, He, and Long]{vcrgt}
Fu, D., Hua, Z., Xie, Y., Fang, J., Zhang, S., Sancak, K., Wu, H., Malevich, A., He, J., and Long, B.
\newblock Vcr-graphormer: {A} mini-batch graph transformer via virtual connections.
\newblock In \emph{Proceedings of the International Conference on Learning Representations}, 2024.

\bibitem[He et~al.(2022{\natexlab{a}})He, Liang, Liu, Wen, Jiao, and Feng]{bmgcn}
He, D., Liang, C., Liu, H., Wen, M., Jiao, P., and Feng, Z.
\newblock Block modeling-guided graph convolutional neural networks.
\newblock In \emph{Proceedings of the Thirty-Sixth {AAAI} Conference on Artificial Intelligence}, 2022{\natexlab{a}}.

\bibitem[He et~al.(2022{\natexlab{b}})He, Chen, Xu, and He]{rlp}
He, Q., Chen, J., Xu, H., and He, K.
\newblock Structural robust label propagation on homogeneous graphs.
\newblock In \emph{Proceedings of the {IEEE} International Conference on Data Mining}, 2022{\natexlab{b}}.

\bibitem[Kipf \& Welling(2017)Kipf and Welling]{gcn}
Kipf, T.~N. and Welling, M.
\newblock {Semi-supervised Classification with Graph Convolutional Networks}.
\newblock In \emph{Proceedings of the International Conference on Learning Representations}, 2017.

\bibitem[Klicpera et~al.(2019)Klicpera, Bojchevski, and G{\"{u}}nnemann]{appnp}
Klicpera, J., Bojchevski, A., and G{\"{u}}nnemann, S.
\newblock Predict then propagate: Graph neural networks meet personalized pagerank.
\newblock In \emph{Proceedings of the International Conference on Learning Representations}, 2019.

\bibitem[Li et~al.(2022)Li, Zhu, Cheng, Shan, Luo, Li, and Qian]{glognn}
Li, X., Zhu, R., Cheng, Y., Shan, C., Luo, S., Li, D., and Qian, W.
\newblock Finding global homophily in graph neural networks when meeting heterophily.
\newblock In \emph{Proceedings of the International Conference on Machine Learning}, 2022.

\bibitem[Luan et~al.(2022)Luan, Hua, Lu, Zhu, Zhao, Zhang, Chang, and Precup]{acmgnn}
Luan, S., Hua, C., Lu, Q., Zhu, J., Zhao, M., Zhang, S., Chang, X., and Precup, D.
\newblock Revisiting heterophily for graph neural networks.
\newblock In \emph{Proceedings of the Annual Conference on Neural Information Processing Systems}, 2022.

\bibitem[Ma et~al.(2024)Ma, He, and Wei]{polyformer}
Ma, J., He, M., and Wei, Z.
\newblock Polyformer: Scalable node-wise filters via polynomial graph transformer.
\newblock In \emph{Proceedings of the {ACM} {SIGKDD} Conference on Knowledge Discovery and Data Mining}, 2024.

\bibitem[Ma et~al.(2023)Ma, Chen, Wu, Song, Wang, and Zheng]{agt}
Ma, X., Chen, Q., Wu, Y., Song, G., Wang, L., and Zheng, B.
\newblock Rethinking structural encodings: Adaptive graph transformer for node classification task.
\newblock In \emph{Proceedings of the {ACM} Web Conference}, 2023.

\bibitem[Meng et~al.(2019)Meng, Liang, Bao, and Zhang]{socialnets}
Meng, Z., Liang, S., Bao, H., and Zhang, X.
\newblock Co-embedding attributed networks.
\newblock In \emph{Proceedings of the Twelfth {ACM} International Conference on Web Search and Data Mining}, 2019.

\bibitem[Pei et~al.(2020)Pei, Wei, Chang, Lei, and Yang]{geomgcn}
Pei, H., Wei, B., Chang, K.~C., Lei, Y., and Yang, B.
\newblock Geom-gcn: Geometric graph convolutional networks.
\newblock In \emph{Proceedings of the International Conference on Learning Representations}, 2020.

\bibitem[Platonov et~al.(2023)Platonov, Kuznedelev, Diskin, Babenko, and Prokhorenkova]{roman}
Platonov, O., Kuznedelev, D., Diskin, M., Babenko, A., and Prokhorenkova, L.
\newblock A critical look at the evaluation of gnns under heterophily: Are we really making progress?
\newblock In \emph{Proceedings of the Eleventh International Conference on Learning Representations}, 2023.

\bibitem[Ramp{\'a}{\v{s}}ek et~al.(2022)Ramp{\'a}{\v{s}}ek, Galkin, Dwivedi, Luu, Wolf, and Beaini]{graphgps}
Ramp{\'a}{\v{s}}ek, L., Galkin, M., Dwivedi, V.~P., Luu, A.~T., Wolf, G., and Beaini, D.
\newblock Recipe for a general, powerful, scalable graph transformer.
\newblock In \emph{Proceedings of the Annual Conference on Neural Information Processing Systems}, 2022.

\bibitem[Vaswani et~al.(2017)Vaswani, Shazeer, Parmar, Uszkoreit, Jones, Gomez, Kaiser, and Polosukhin]{transformer}
Vaswani, A., Shazeer, N., Parmar, N., Uszkoreit, J., Jones, L., Gomez, A.~N., Kaiser, {\L}., and Polosukhin, I.
\newblock {Attention Is All You Need}.
\newblock In \emph{Proceedings of the Annual Conference on Neural Information Processing Systems}, 2017.

\bibitem[Veli{\v{c}}kovi{\'c} et~al.(2018)Veli{\v{c}}kovi{\'c}, Cucurull, Casanova, Romero, Lio, and Bengio]{gat}
Veli{\v{c}}kovi{\'c}, P., Cucurull, G., Casanova, A., Romero, A., Lio, P., and Bengio, Y.
\newblock {Graph Attention Networks}.
\newblock In \emph{Proceedings of the International Conference on Learning Representations}, 2018.

\bibitem[Wang et~al.(2019)Wang, Ji, Shi, Wang, Ye, Cui, and Yu]{acm}
Wang, X., Ji, H., Shi, C., Wang, B., Ye, Y., Cui, P., and Yu, P.~S.
\newblock Heterogeneous graph attention network.
\newblock In \emph{Proceedings of the World Wide Web Conference}, 2019.

\bibitem[Wang et~al.(2020)Wang, Zhu, Bo, Cui, Shi, and Pei]{amgcn}
Wang, X., Zhu, M., Bo, D., Cui, P., Shi, C., and Pei, J.
\newblock {AM-GCN:} adaptive multi-channel graph convolutional networks.
\newblock In \emph{Proceedings of the {ACM} {SIGKDD} Conference on Knowledge Discovery and Data Mining}, 2020.

\bibitem[Wu et~al.(2019)Wu, Souza, Zhang, Fifty, Yu, and Weinberger]{sgc}
Wu, F., Souza, A., Zhang, T., Fifty, C., Yu, T., and Weinberger, K.
\newblock {Simplifying Graph Convolutional Networks}.
\newblock In \emph{Proceedings of the International Conference on Machine Learning}, 2019.

\bibitem[Wu et~al.(2022)Wu, Zhao, Li, Wipf, and Yan]{nodeformer}
Wu, Q., Zhao, W., Li, Z., Wipf, D., and Yan, J.
\newblock Nodeformer: A scalable graph structure learning transformer for node classification.
\newblock In \emph{Proceedings of the Annual Conference on Neural Information Processing Systems}, 2022.

\bibitem[Wu et~al.(2023)Wu, Zhao, Yang, Zhang, Nie, Jiang, Bian, and Yan]{sgformer}
Wu, Q., Zhao, W., Yang, C., Zhang, H., Nie, F., Jiang, H., Bian, Y., and Yan, J.
\newblock Simplifying and empowering transformers for large-graph representations.
\newblock In \emph{Proceedings of the Annual Conference on Neural Information Processing Systems}, 2023.

\bibitem[Xing et~al.(2024)Xing, Wang, Li, Huang, and Shi]{cob}
Xing, Y., Wang, X., Li, Y., Huang, H., and Shi, C.
\newblock Less is more: on the over-globalizing problem in graph transformers.
\newblock In \emph{Proceedings of the International Conference on Machine Learning}, 2024.

\bibitem[Xu et~al.(2018)Xu, Li, Tian, Sonobe, Kawarabayashi, and Jegelka]{jknet}
Xu, K., Li, C., Tian, Y., Sonobe, T., Kawarabayashi, K.-i., and Jegelka, S.
\newblock Representation learning on graphs with jumping knowledge networks.
\newblock In \emph{Proceedings of the International conference on machine learning}, 2018.

\bibitem[Zhang et~al.(2023)Zhang, Cheng, and Zhang]{gnn2}
Zhang, G., Cheng, D., and Zhang, S.
\newblock {FPGNN:} fair path graph neural network for mitigating discrimination.
\newblock \emph{World Wide Web}, 26\penalty0 (5):\penalty0 3119--3136, 2023.

\bibitem[Zhang et~al.(2024)Zhang, Cheng, Yuan, and Zhang]{gnn1}
Zhang, G., Cheng, D., Yuan, G., and Zhang, S.
\newblock Learning fair representations via rebalancing graph structure.
\newblock \emph{Inf. Process. Manag.}, 61\penalty0 (1):\penalty0 103570, 2024.

\bibitem[Zhang et~al.(2025)Zhang, Yuan, Cheng, Liu, Li, and Zhang]{gnn3}
Zhang, G., Yuan, G., Cheng, D., Liu, L., Li, J., and Zhang, S.
\newblock Disentangled contrastive learning for fair graph representations.
\newblock \emph{Neural Networks}, 181:\penalty0 106781, 2025.

\bibitem[Zhang et~al.(2022)Zhang, Liu, Hu, and Lee]{ansgt}
Zhang, Z., Liu, Q., Hu, Q., and Lee, C.
\newblock {Hierarchical Graph Transformer with Adaptive Node Sampling}.
\newblock In \emph{Proceedings of the Annual Conference on Neural Information Processing Systems}, 2022.

\bibitem[Zhao et~al.(2021)Zhao, Li, Wen, Wang, Liu, Sun, Xie, and Ye]{gophormer}
Zhao, J., Li, C., Wen, Q., Wang, Y., Liu, Y., Sun, H., Xie, X., and Ye, Y.
\newblock {Gophormer: Ego-Graph Transformer for Node Classification}.
\newblock \emph{arXiv preprint arXiv:2110.13094}, 2021.

\bibitem[Zhu et~al.(2020)Zhu, Yan, Zhao, Heimann, Akoglu, and Koutra]{h2gnn}
Zhu, J., Yan, Y., Zhao, L., Heimann, M., Akoglu, L., and Koutra, D.
\newblock Beyond homophily in graph neural networks: Current limitations and effective designs.
\newblock In \emph{Proceedings of the Annual Conference on Neural Information Processing Systems}, 2020.

\end{thebibliography}
